\documentclass[conference]{IEEEtran}
\IEEEoverridecommandlockouts

\usepackage{amsmath,graphicx, multirow}

\usepackage{amsmath,graphicx, multirow,subcaption, float, hyperref}

\usepackage{cite}
\usepackage{amsmath,amssymb,amsfonts}
\usepackage{algorithmic}
\usepackage{graphicx}
\usepackage{textcomp}
\usepackage{xcolor}
\def\BibTeX{{\rm B\kern-.05em{\sc i\kern-.025em b}\kern-.08em
    T\kern-.1667em\lower.7ex\hbox{E}\kern-.125emX}}
\begin{document}

\title{Efficacy of Synthetic Data as a Benchmark}

\author{
    \IEEEauthorblockN{
        Gaurav Maheshwari\IEEEauthorrefmark{2}, Dmitry Ivanov\IEEEauthorrefmark{2}, Kevin El Haddad\IEEEauthorrefmark{2}\IEEEauthorrefmark{3}
    }
    \IEEEauthorblockA{\IEEEauthorrefmark{2} Diabolocom}
    \IEEEauthorblockA{\IEEEauthorrefmark{3} ISIA Lab - University of Mons}
    \IEEEauthorblockA{first\_name.last\_name@diabolocom.com}
}

\maketitle

\begin{abstract}
Large language models (LLMs) have enabled a range of applications in zero-shot and few-shot learning settings, including the generation of synthetic datasets for training and testing. However, to reliably use these synthetic datasets, it is essential to understand how representative they are of real-world data. We investigate this by assessing the effectiveness of generating synthetic data through LLM and using it as a benchmark for various NLP tasks. Our experiments across six datasets, and three different tasks, show that while synthetic data can effectively capture performance of various methods for simpler tasks, such as intent classification, it falls short for more complex tasks like named entity recognition. Additionally, we propose a new metric called the bias factor, which evaluates the biases introduced when the same LLM is used to both generate benchmarking data and to perform the tasks. We find that smaller LLMs exhibit biases towards their own generated data, whereas larger models do not. Overall, our findings suggest that the effectiveness of synthetic data as a benchmark varies depending on the task, and that practitioners should rely on data generated from multiple larger models whenever possible.
\end{abstract}

\begin{IEEEkeywords}
Synthetic data, LLM, NLP, Benchmark
\end{IEEEkeywords}

\section{Introduction}
\label{sec:intro}

In many real-world machine learning applications, obtaining and annotating large datasets is costly or impractical. Furthermore, many industrial domains such as healthcare~\cite{gostin2009beyond, qayyum2020secure} and banking~\cite{leo2019machine, culnan2003consumer} impose stringent privacy constraints, preventing the use of annotated data for training or evaluation. In response, large language models (LLMs) have emerged as a promising solution, demonstrating exceptional capabilities in few-shot and zero-shot learning settings~\cite{kojima2022large, monazzah2024zero, guo2023images, li2024flexkbqa}. Consequently, a variety of LLMs, both open-source and proprietary, are available, each with distinct performance characteristics. However, from the perspective of the practitioner, with limited or no access to test data, determining the most suitable LLM for a specific task remains a challenging and often unclear decision.

To address this challenge, our work explores the use of synthetic data generated by LLMs as a benchmark for assessing model performance. Unlike previous research~\cite{nowruzi2019much, jaipuria2020deflating, tang2023does}, which primarily focused on improving results by combining synthetic and real data, our study investigates the extent to which synthetic data can predict the zero-shot performance of LLMs on various NLP tasks.

To this end, we generate synthetic data using LLMs across six datasets and three tasks: intent detection, text similarity, and named entity recognition (NER). We evaluate the effectiveness of synthetic data as a benchmark by analyzing two key aspects: (i) absolute performance, which measures the difference in the performance of the model on real versus synthetic datasets, and (ii) relative performance, which assesses changes in the relative ranking of methods when evaluated on real versus synthetic datasets. Additionally, we explore the potential bias that may arise when the same LLM is used both to generate synthetic data and to solve the task. To quantify this bias, we propose a new metric called the \textit{bias factor}, which takes into account the performance of LLM on its own synthetic data relative to its performance on data generated by other LLMs.

Our experiments reveal that synthetic data is particularly effective as a benchmark for simpler tasks like intent detection, but its representativeness diminishes for more complex tasks like NER. We do not find any particular LLM significantly better at producing synthetic data than others, as different LLMs perform best for different tasks. Moreover, we observe that averaging performance across synthetic data generated by multiple LLMs results in a more robust and representative benchmark. Interestingly, larger LLMs do not exhibit significant bias, while smaller models tend to perform slightly better on datasets they generated themselves.

In summary, our main contributions and conclusions are:

\begin{itemize}
    \item We systematically evaluate the efficacy of generating data via LLMs and using it as a benchmark to predict the performance of various methods. We find that synthetic data is particularly representative for simpler tasks like intent detection, while its effectiveness diminishes for more complex tasks.
    \item We introduce the \textit{bias factor} metric to quantify potential bias when the same LLM is used for both data generation and task solving. Our results show that larger models exhibit little to no bias, whereas smaller models with fewer parameters tend to perform better on their own data.
\end{itemize}

\section{Setup and Evaluation Protocol}
\label{sec:approach}

In this section, we begin by illustrating our synthetic data generation mechanism. We then describe our two evaluation protocols designed to assess the effectiveness of synthetic data as a benchmark as well as the bias introduced by using the same LLM for generation and using it to solve the task.

\subsection{Data Generation and Prediction}
\label{subsec:data_generation_and_prediction}


In this work, we employ a prompt-based approach for both generating synthetic data via LLM and using it as a method to solve the task at hand. For any given task, following the approach of~\cite{long2024llms}, we structure the data generation prompt into four main components:

\begin{itemize}
    \item Task description: A description of the task along with annotation guidelines, if available in the original dataset;
    \item Examples: A few examples sampled from the real dataset. For the intent classification and text similarity tasks, we sample $k$ examples for each intent and score, respectively. In the case of NER, we randomly sample $k'$ examples from the dataset.
    \item Generation guidelines: Specific characteristics of the synthetic data, such as length, tone, or other relevant attributes;
    \item Output description: Details the desired output format
\end{itemize}


When using LLM to solve tasks in a zero-shot setting, we follow a similar prompt-based approach. The prompt is composed of three main elements: (i) Task description, which is similar to the data generation prompt; (ii) Output description, which specifies the output format; and (iii) the input data point for which the output needs to be predicted. The exact prompts for both data generation and task solving are available in the GitHub repository\footnote{\url{https://github.com/Diabolocom-Research/SG4NLP}}.

\subsection{Efficacy of Synthetic Data}
\label{subsec:efficacy}

The primary objective of this study is to assess whether evaluating various methods using synthetic data yields comparable performance to real data benchmarks. Specifically, we aim to determine if different methods produce the same performance metrics when tested on synthetic versus real datasets. However, even if there are substantial differences in absolute performance, synthetic data can still be valuable if it preserves the relative ranking of methods as seen on real data. This ranking similarity would allow practitioners to identify the most effective approach for a given task. To capture these two aspects, we define and measure absolute and relative performance in our evaluation.

\begin{table*}[t!]
    \centering
        \footnotesize

    \begin{subtable}{\linewidth}
        \centering
            \begin{tabular}{llllllllllll}
                \hline
                \multicolumn{1}{c}{\multirow{3}{*}{LLM}} &
                  \multicolumn{5}{c}{Headlines} &
                  \multicolumn{1}{c}{} &
                  \multicolumn{5}{c}{Tweet-News} \\ \cline{2-6} \cline{8-12} 
                \multicolumn{1}{c}{} &
                  \multicolumn{2}{c}{Pearson} &
                  \multicolumn{1}{c}{} &
                  \multicolumn{2}{c}{Spearman} &
                  \multicolumn{1}{c}{} &
                  \multicolumn{2}{c}{Pearson} &
                  \multicolumn{1}{c}{} &
                  \multicolumn{2}{c}{Spearman} \\ \cline{2-3} \cline{5-6} \cline{8-9} \cline{11-12} 
                \multicolumn{1}{c}{}            & MSPD & SRCC   &  & MSPD & SRCC   &  & MSPD & SRCC   &  & MSPD & SRCC \\ \hline
                        GPT-4o                  & 0.07 & 0.83 &  & 0.07 & 0.83 &  & 0.09 & 0.83 &  & 0.12 & 0.94  \\
                        GPT-4o-mini             & 0.06 & 0.77 &  & 0.06 & 0.77 &  & 0.08 & 0.83 &  & 0.12 & 0.77  \\
                        Llama-3-70B             & 0.05 & 1.0  &  & 0.05 & 0.77 &  & 0.03 & 0.94 &  & 0.04 & 1.0  \\
                        Llama-3-8B              & 0.02 & 0.77 &  & 0.02 & 0.77 &  & 0.03 & 0.77 &  & 0.03 & 0.54   \\
                        Mixtral-8x22B           & 0.05 & 0.83 &  & 0.04 & 0.83 &  & 0.06 & 0.94 &  & 0.09 & 0.77  \\
                        Mixtral-8x7B            & 0.03 & 0.77 &  & 0.03 & 0.77 &  & 0.03 & 0.77 &  & 0.03 & 0.94  \\
                        Average                 & 0.04 & 0.83 &  & 0.04 & 0.83 &  & 0.03 & 0.83 &  & 0.05 & 0.94  \\ \hline
            \end{tabular}
        \caption{Results over Headlines and Tweet-news (text similarity task) with 1 example per score}
        \end{subtable} \par

        \begin{subtable}{\linewidth}
            \centering
            \begin{tabular}{llllllllllllllllll}
            \hline
            \multirow{3}{*}{LLM} & \multicolumn{8}{c}{Literature}                &  & \multicolumn{8}{c}{Politics}                  \\ \cline{2-9} \cline{11-18} 
             &
              \multicolumn{2}{c}{strict F1} &
               &
              \multicolumn{2}{c}{partial F1} &
               &
              \multicolumn{2}{c}{entity type} &
               &
              \multicolumn{2}{c}{strict F1} &
               &
              \multicolumn{2}{c}{partial F1} &
               &
              \multicolumn{2}{c}{entity type} \\ \cline{2-3} \cline{5-6} \cline{8-9} \cline{11-12} \cline{14-15} \cline{17-18} 
                                    & MSPD   & SRCC   &  & MSPD   & SRCC   &  & MSPD   & SRCC   &  & MSPD   & SRCC   &  & MSPD   & SRCC   &  & MSPD   & SRCC   \\ \hline
            GPT-4o                  & 0.02 & 0.77 &  & 0.01 & 0.60 &  & 0.01 & 0.77 &  & 0.01 & 0.71 &  & 0.01 & 0.49 &  & 0.01 & 0.83 \\
            GPT-4o-mini             & 0.01 & 0.83 &  & 0.01 & 0.60 &  & 0.01 & 0.83 &  & 0.01 & 0.60 &  & 0.01 & 0.09 &  & 0.02 & 0.66 \\
            Llama-3-70B             & 0.01 & 0.77 &  & 0.00 & 0.66 &  & 0.00 & 0.77 &  & 0.01 & 0.83 &  & 0.02 & 0.71 &  & 0.02 & 0.94 \\
            Llama-3-8B              & 0.00 & 0.94 &  & 0.02 & 0.54 &  & 0.00 & 0.83 &  & 0.03 & 0.43 &  & 0.05 & 0.20 &  & 0.03 & 0.60 \\
            Mixtral-8x22B           & 0.01 & 0.60 &  & 0.00 & 0.60 &  & 0.01 & 0.49 &  & 0.02 & 0.77 &  & 0.02 & 0.60 &  & 0.02 & 1.0  \\
            Mixtral-8x7B            & 0.00 & 0.83 &  & 0.01 & 0.43 &  & 0.00 & 0.66 &  & 0.02 & 0.89 &  & 0.02 & 0.14 &  & 0.02 & 0.94 \\
            Average                 & 0.00 & 0.77 &  & 0.00 & 0.60 &  & 0.00 & 0.77 &  & 0.02 & 0.71 &  & 0.02 & 0.20 &  & 0.02 & 0.94 \\
            \hline
            \end{tabular}
            \caption{Results over CrossNER-Literature and CrossNER-Politics (NER task) with 5 examples.}
            \end{subtable} \par

            \begin{subtable}{\linewidth}
                \centering
                \begin{tabular}{llllllllllllllllll}
                \hline
                \multirow{3}{*}{LLM} & \multicolumn{8}{c}{SNIPS}                     &  & \multicolumn{8}{c}{ATIS}                      \\ \cline{2-9} \cline{11-18} 
                 &
                  \multicolumn{2}{c}{Precision} &
                   &
                  \multicolumn{2}{c}{Recall} &
                   &
                  \multicolumn{2}{c}{F1} &
                   &
                  \multicolumn{2}{c}{Precision} &
                   &
                  \multicolumn{2}{c}{Recall} &
                   &
                  \multicolumn{2}{c}{F1} \\ \cline{2-3} \cline{5-6} \cline{8-9} \cline{11-12} \cline{14-15} \cline{17-18} 
                                        & MSPD   & SRCC   &  & MSPD   & SRCC   &  & MSPD    & SRCC   &  & MSPD   & SRCC   &  & MSPD   & SRCC   &  & MSPD   & SRCC   \\ \hline
                GPT-4o                  & 0.00 & 0.94 &  & 0.00 & 0.94 &  & 0.00  & 0.94 &  & 0.11 & 0.89 &  & 0.09 & 0.89 &  & 0.11 & 0.94 \\
                GPT-4o-mini             & 0.00 & 0.84 &  & 0.00  & 0.90 &  & 0.00 & 0.90 &  & 0.13 & 0.89 &  & 0.09 & 0.89 &  & 0.12 & 0.94 \\
                Llama-3-70B             & 0.00 & 0.94 &  & 0.00  & 0.83 &  & 0.00 & 0.83 &  & 0.06 & 0.71 &  & 0.03 & 0.89 &  & 0.04 & 0.94 \\
                Llama-3-8B              & 0.00 & 0.94 &  & 0.00  & 0.94 &  & 0.00 & 0.94 &  & 0.08 & 0.60 &  & 0.04 & 0.71 &  & 0.05 & 0.77 \\
                Mixtral-8x22B           & 0.00 & 0.94 &  & 0.00 & 0.99 &  & 0.01  & 0.94 &  & 0.07 & 0.89 &  & 0.03 & 0.89 &  & 0.05 & 0.94 \\
                Mixtral-8x7B            & 0.00 & 0.89 &  & 0.00 & 0.94 &  & 0.00  & 0.94 &  & 0.1  & 0.60 &  & 0.05 & 0.89 &  & 0.07 & 0.94 \\
                Average                 & 0.00 & 0.94 &  & 0.00 & 0.94 &  & 0.00  & 0.94 &  & 0.08 & 0.89 &  & 0.05 & 0.89 &  & 0.06 & 0.94 \\
                \hline
                \end{tabular}
                \caption{Results over SNIPS and ATIS (intent classification task) with 1 example per intent}
                \end{subtable}
\caption{Results on evaluating the efficacy of synthetic data as a substitute for real data for benchmarking. We report using Mean Square Performance Difference (MSPD), as described in Section~\ref{subsec:efficacy}, where lower values indicate better absolute performance. To capture relative performance, we use Spearman rank correlation coefficient (SRCC), where higher values indicate better performance. Apart from all LLMs, we also report Average which is the mean of performance metric of individual LLMs.}
\label{tab:exp1}
\end{table*}


To formally define these notions, consider a set of $n$ different approaches $\{\mathbf{A}_1, \cdots, \mathbf{A}_n\}$ and a dataset $\mathcal{D}$. Note that, in our study we use LLM as the data generation mechanism as well as the approach to solve the task. Let $\textit{m}$ be the performance metric (such as accuracy or F1 measure), with performance values for approaches ${\mathbf{A}_1, \cdots, \mathbf{A}_n}$ as $\textit{m}^{\mathbf{A}_1}_{\mathcal{D}}, \cdots, \textit{m}^{\mathbf{A}_n}_{\mathcal{D}}$ respectively. Additionally, consider a synthetic dataset $\mathcal{S}$, generated by some LLM, by using examples for the dataset $\mathcal{D}$. Let the performance of the aforementioned approaches on $\mathcal{S}$ be $\textit{m}^{\mathbf{A}_1}_{\mathcal{S}}, \cdots, \textit{m}^{\mathbf{A}_n}_{\mathcal{S}}$. We hypothesize that the synthetic data is representative if:
\begin{itemize}
    \item Absolute performance: $\textit{m}^{\mathbf{A}_i}_{\mathcal{D}}$ is similar to $\textit{m}^{\mathbf{A}_i}_{\mathcal{S}}$ for all $i$

    \item Relative performance: High correlation between the ranks of $\textit{m}^{\mathbf{A}_1}_{\mathcal{D}}, \cdots, \textit{m}^{\mathbf{A}_n}_{\mathcal{D}}$ and $\textit{m}^{\mathbf{A}_1}_{\mathcal{S}}, \cdots, \textit{m}^{\mathbf{A}_n}_{\mathcal{S}}$. 
\end{itemize}

To capture the absolute performance difference, we propose mean squared performance difference (MSPD):

\begin{equation}
    MSPD(\textit{m}, \mathcal{D}, \mathcal{S}) = \frac{\sum_{i \in \{\mathbf{A}_1, \cdots, \mathbf{A}_n\}} (\textit{m}^{\mathbf{A}_i}_{\mathcal{D}} - \textit{m}^{\mathbf{A}_i}_{\mathcal{S}})^2}{|\mathbf{A}_1, \cdots, \mathbf{A}_n|}
\end{equation}

MSPD, effectively captures the average difference in the performance of approaches on the real dataset and the synthetic dataset under consideration. However, even with high MSPD, the synthetic dataset could be useful as the rank of different approaches on the real and synthetic dataset might be similar. 
To capture this relative performance, we rely on Spearman's rank correlation~\cite{spearman1961proof} measure between the two sets $\textit{m}^{\mathbf{A}_1}_{\mathcal{D}}, \cdots, \textit{m}^{\mathbf{A}_n}_{\mathcal{D}}$ and $\textit{m}^{\mathbf{A}_1}_{\mathcal{S}}, \cdots, \textit{m}^{\mathbf{A}_n}_{\mathcal{S}}$.

A high Spearman's correlation implies that the ranking of approaches over the real dataset is the same as over the synthetic dataset, indicating that the synthetic data accurately captures relative performance. Conversely, a high MSPD implies that the methods have dissimilar absolute performance on the synthetic and real data.

\subsection{Bias Factor}
\label{subsec:bias_factor}

To evaluate the  bias introduced by using the same LLM for generating synthetic data and solving the task, a straightforward approach is to directly compare the performance of the LLM on its own data with its performance on data generated by other LLMs. However, our experiments revealed substantial performance differences when using data from different LLMs. In other words, we found high MSPD, as defined in the subsection above. To account for these discrepancies, we first normalize the performances and then compare them.

More specifically, consider a set of LLMs $\{A, \cdots, P\}$, a dataset $\mathcal{D}$, and a performance measure $\textit{m}$. Let $\mathcal{D}^K$ denote the dataset generated by LLM $K$ using examples from $\mathcal{D}$. Let $\textit{m}^K_{\mathcal{D}}$ represent the performance of LLM $K$ on $\mathcal{D}$. We define normalized performance $\textit{nm}^K_{\mathcal{D}^K}$ as:

\begin{equation}
     \textit{nm}^K_{\mathcal{D}^K} =  \textit{m}^K_{\mathcal{D}^K} - \frac{\sum_{i \in {A,\cdots,P} }^{} \textit{m}^i_{\mathcal{D}^K} }{|{A, \cdots, P}|}
     \label{eq:normalized_performance}
\end{equation}

Finally, the bias factor for LLM $K$ for dataset $\mathcal{D}$ is:
\begin{equation}
     bf =  \textit{nm}^K_{\mathcal{D}^K} - \frac{\sum_{i \in {A,\cdots, J, L, \cdots P} }^{} \textit{nm}^K_{\mathcal{D}^i} }{|{A,\cdots, J, L, \cdots P}|}
     \label{eq:comparison_equation}
\end{equation}



In equation~\ref{eq:normalized_performance}, we normalize the performance of LLM $K$ by subtracting the average performance of all approaches on the dataset generated by $K$. Subsequently, in equation~\ref{eq:comparison_equation}, we compare the normalized performance of LLM $K$ on its own dataset to its average normalized performance across datasets generated by other LLMs.

For a performance measure between 0 and 1, the range of the bias factor is between -1 and 1, with a positive value indicating a bias towards its own data. For performance measures greater than 1, we recommend normalizing them to fall between 0 and 1.

\begin{table*}[t!]
\small
\centering
\begin{tabular}{llllllllll}
\hline
\multicolumn{1}{c}{\multirow{2}{*}{Methods}} &
  \multicolumn{2}{c}{NER} &
  \multicolumn{1}{c}{} &
  \multicolumn{2}{c}{Intent} &
  \multicolumn{1}{c}{} &
  \multicolumn{2}{c}{Text Similarity} &
  \multicolumn{1}{c}{\multirow{2}{*}{Average}} \\ \cline{2-3} \cline{5-6} \cline{8-9}
\multicolumn{1}{c}{} & Literature & Politics &  & Snips  & Atis   &  & Headlines & Tweet-News & \multicolumn{1}{c}{} \\ \hline
GPT-4o               & +0.045      & +0.062    &  & -0.02  & +0.113  &  & +0.007     & -0.009     & +0.033                \\
GPT-4o-mini          & -0.014     & -0.055   &  & -0.014 & +0.137  &  & -0.005    & +0.07       & +0.019                \\
Llama-3-70B          & +0.012      & -0.023   &  & +0.001  & +0.007  &  & -0.018    & -0.074     & -0.095               \\
Llama-3-8B           & +0.004      & -0.023   &  & +0.009  & -0.019 &  & -0.079    & -0.102     & -0.038               \\
Mixtral-8x22B        & +0.017      & +0.002    &  & +0.016  & +0.058  &  & +0.016     & -0.012     & +0.016                \\
Mixtral-8x7B         & +0.041      & +0.04     &  & +0.004  & +0.017  &  & +0.047     & +0.223      & +0.061    \\
\hline
\end{tabular}
\caption{Bias factor of multiple LLMs across various datasets. For NER, we use the Exact F1 score as the performance metric. While for intent recognition, we use the average F1 score and for the text similarity task, we use the Pearson correlation coefficient. The average is the mean bias factor across all the task.~\label{tab:bias-factor}}
\end{table*}

\section{Experiments}

In this section, we present experiments on: (i) the effectiveness of synthetic data as a benchmark, and (ii) the bias introduced when using the same LLM for both generating the dataset and applying it to solve the task. We begin by outlining the datasets and experimental setup, followed by presenting and discussing the results.

\textbf{Task and Datasets:} We conduct our experiments on three different tasks, each with two datasets, as follows:

\begin{itemize}
    \item \textbf{Intent Detection} is a classification task that involves determining the underlying goal behind a user query. In this study, we use: (a) SNIPS~\cite{coucke2018snips}—a collection of utterances with 7 associated intents, and (b) Airline Travel Information System (ATIS)—a dataset containing manual transcripts of flight information with 17 intents.
    \item \textbf{Named Entity Recognition} tasks involve extracting information such as names and places from user text. This generally involves first identifying the span of the entity in the text and then predicting its type. In this work, we use two variants of Cross-NER~\cite{liu2020crossner}: (a) Literature, which includes texts containing entities related to historical literature, and (b) Politics, which features entities associated with political parties.
    \item \textbf{Text Similarity} involves assigning a similarity score between two given sentences. In this work, we use the SemEval 2012 dataset~\cite{agirre2012semeval}, which consists of text pairs with associated scores, focusing on two domains: (a) Headlines, consisting of pairs of news headlines, and (b) Tweets-News, consisting of tweet pairs.
\end{itemize}

For each of the dataset we sample 200 examples for testing.

\textbf{Metrics:} For the intent recognition task, we report the average macro F1 score, along with precision and recall. For text similarity, we follow~\cite{agirre2012semeval} and report Pearson and Spearman's rank correlation coefficients. For NER, we report the F1 score for both exact and partial matches of entity spans, as well as the F1 scores for entity type prediction.

\textbf{Models:} In our experiments, we focus on three major families of models composed of proprietary and open source models, namely: (i) GPT, which includes GPT-4o and GPT-4o-mini; (ii) Mixtral, consisting of Mixtral-8x22B-Instruct-v0.1 and Mixtral-8x7B-Instruct-v0.1; and (iii) Llama, comprising Meta-Llama-3-70B-Instruct and Meta-Llama-3-8B-Instruct.

\subsection{Efficacy of Synthetic Data}

In this experiment, we aim to evaluate the use of synthetic data generated by LLMs (see Section~\ref{subsec:data_generation_and_prediction}) as a replacement for real data in benchmarking various methods. As detailed in Section~\ref{subsec:efficacy}, we report the Mean Square Performance Difference (MSPD) to compare absolute performance and the Spearman's rank correlation coefficient (SRCC) to evaluate relative performance. A low MSPD indicates that the generated data accurately captures absolute performance, whereas a high MSPD suggests it does not. Conversely, a high SRCC implies that synthetic data accurately captures relative performance, while a low SRCC indicates otherwise. 

The results for text similarity, NER, and intent classification are presented in Table~\ref{tab:exp1} a, b, and c, respectively. For text similarity, we find that synthetic data is not representative of the absolute performance, as most LLMs show high MSPD. However, synthetic data can capture relative performance, as all LLMs exhibit high SRCC, with Llama-3-70B showing a correlation of 1.0 over few performance metrics. Similar results are observed in intent classification, where all LLMs show high SRCC. In SNIPS, however, unlike ATIS, we find low MPSD, indicating that the synthetic data captures both relative and absolute performance. We attribute this exception to the relatively simpler dataset with a limited number of intents ($7$ in SNIPS, while $17$ in ATIS). For NER, we observe that SRCC significantly decreases, particularly for partial F1 and entity type prediction. We attribute this lower SRCC to the fact that NER is a relatively more complex task, involving multiple span prediction, unlike other tasks that generally involve only one prediction step.

In summary, we find synthetic data to be a good predictor of relative performance but not a strong indicator of absolute performance. Furthermore, we do not find synthetic data from any one LLM to be more representative than that from others. Each LLM has strengths—datasets and metrics where they generate representative data—and weaknesses—datasets and metrics where they are suboptimal. Overall, we recommend practitioners to generate data from multiple LLMs and average the benchmarks for more robust and representative results.

\subsection{Bias Factor}

In the previous experiment, we explored the role of synthetic data in benchmarking. Here, we take a closer look at the bias introduced by using the same LLM for both generating synthetic data and solving the task. To this end, we follow the protocol described in Section~\ref{subsec:bias_factor} and report bias factor.

The results of this experiment are presented in Table~\ref{tab:bias-factor}. We find that for LLama and Mixtral, smaller LLMs show stronger bias compared to larger LLMs. In particular, Mixtral-8x7B exhibits a strong bias towards its own data, with an average bias factor of 0.061. However, the trend is reversed in the case of the GPT family, where GPT-4o shows a higher bias factor than GPT-4o-mini. From the perspective of tasks, we find that LLMs show lower bias factors for SNIPS and headlines, while higher average bias factors are observed for NER tasks, as well as ATIS and Tweet-News.

Overall, we find that smaller LLMs generally exhibit slightly higher bias factors than larger LLMs. Additionally, our observations suggest that the nature of the task itself influences the bias exhibited by LLMs, with easier tasks showing lower bias and more complex tasks showing higher bias.



\section{Conclusion and Perspective}

In this work, we evaluate the use of synthetic data as a substitute for real data in benchmarking. Our experiments reveal that for easier tasks like intent detection, synthetic data can effectively predict the performance of various approaches. However, for more complex tasks like NER, this predictive capability diminishes. Additionally, we propose a new evaluation metric called the bias factor, which measures the potential bias introduced when the same LLM is used for both generating data and solving tasks. We observe that smaller models tend to perform better on data they generated themselves.

One limitation of our study is the use of the same prompt across all LLMs, which may cause certain models to underperform. Additionally, the evaluation is limited to a zero-shot setting. In the future, we aim to explore few-shot settings, where models are provided with a small number of examples. We also plan to expand our study by including more tasks and additional LLMs.

\bibliographystyle{IEEEbib}
\bibliography{refs}

\begin{thebibliography}{10}

\bibitem{gostin2009beyond}
Lawrence~O Gostin, Laura~A Levit, and Sharyl~J Nass,
\newblock ``Beyond the hipaa privacy rule: enhancing privacy, improving health through research,''
\newblock 2009.

\bibitem{qayyum2020secure}
Adnan Qayyum, Junaid Qadir, Muhammad Bilal, and Ala Al-Fuqaha,
\newblock ``Secure and robust machine learning for healthcare: A survey,''
\newblock {\em IEEE Reviews in Biomedical Engineering}, vol. 14, pp. 156--180, 2020.

\bibitem{leo2019machine}
Martin Leo, Suneel Sharma, and Koilakuntla Maddulety,
\newblock ``Machine learning in banking risk management: A literature review,''
\newblock {\em Risks}, vol. 7, no. 1, pp. 29, 2019.

\bibitem{culnan2003consumer}
Mary~J Culnan and Robert~J Bies,
\newblock ``Consumer privacy: Balancing economic and justice considerations,''
\newblock {\em Journal of social issues}, vol. 59, no. 2, pp. 323--342, 2003.

\bibitem{kojima2022large}
Takeshi Kojima, Shixiang~Shane Gu, Machel Reid, Yutaka Matsuo, and Yusuke Iwasawa,
\newblock ``Large language models are zero-shot reasoners,''
\newblock {\em Advances in neural information processing systems}, vol. 35, pp. 22199--22213, 2022.

\bibitem{monazzah2024zero}
Erfan~Moosavi Monazzah and Mahdi Feghhi,
\newblock ``Zero shot is all you need at semeval-2024 task 9: A study of state of the art llms on lateral thinking puzzles,''
\newblock in {\em Proceedings of the 18th International Workshop on Semantic Evaluation (SemEval-2024)}, 2024, pp. 1889--1893.

\bibitem{guo2023images}
Jiaxian Guo, Junnan Li, Dongxu Li, Anthony Meng~Huat Tiong, Boyang Li, Dacheng Tao, and Steven Hoi,
\newblock ``From images to textual prompts: Zero-shot visual question answering with frozen large language models,''
\newblock in {\em Proceedings of the IEEE/CVF conference on computer vision and pattern recognition}, 2023, pp. 10867--10877.

\bibitem{li2024flexkbqa}
Zhenyu Li, Sunqi Fan, Yu~Gu, Xiuxing Li, Zhichao Duan, Bowen Dong, Ning Liu, and Jianyong Wang,
\newblock ``Flexkbqa: A flexible llm-powered framework for few-shot knowledge base question answering,''
\newblock in {\em Proceedings of the AAAI Conference on Artificial Intelligence}, 2024, vol.~38, pp. 18608--18616.

\bibitem{nowruzi2019much}
Farzan~Erlik Nowruzi, Prince Kapoor, Dhanvin Kolhatkar, Fahed~Al Hassanat, Robert Laganiere, and Julien Rebut,
\newblock ``How much real data do we actually need: Analyzing object detection performance using synthetic and real data,''
\newblock {\em arXiv preprint arXiv:1907.07061}, 2019.

\bibitem{jaipuria2020deflating}
Nikita Jaipuria, Xianling Zhang, Rohan Bhasin, Mayar Arafa, Punarjay Chakravarty, Shubham Shrivastava, Sagar Manglani, and Vidya~N Murali,
\newblock ``Deflating dataset bias using synthetic data augmentation,''
\newblock in {\em Proceedings of the IEEE/CVF Conference on Computer Vision and Pattern Recognition Workshops}, 2020, pp. 772--773.

\bibitem{tang2023does}
Ruixiang Tang, Xiaotian Han, Xiaoqian Jiang, and Xia Hu,
\newblock ``Does synthetic data generation of llms help clinical text mining?,''
\newblock {\em arXiv preprint arXiv:2303.04360}, 2023.

\bibitem{long2024llms}
Lin Long, Rui Wang, Ruixuan Xiao, Junbo Zhao, Xiao Ding, Gang Chen, and Haobo Wang,
\newblock ``On llms-driven synthetic data generation, curation, and evaluation: A survey,''
\newblock {\em arXiv preprint arXiv:2406.15126}, 2024.

\bibitem{spearman1961proof}
Charles Spearman,
\newblock ``The proof and measurement of association between two things.,''
\newblock 1961.

\bibitem{coucke2018snips}
Alice Coucke, Alaa Saade, Adrien Ball, Th{\'e}odore Bluche, Alexandre Caulier, David Leroy, Cl{\'e}ment Doumouro, Thibault Gisselbrecht, Francesco Caltagirone, Thibaut Lavril, et~al.,
\newblock ``Snips voice platform: an embedded spoken language understanding system for private-by-design voice interfaces,''
\newblock {\em arXiv preprint arXiv:1805.10190}, 2018.

\bibitem{liu2020crossner}
Zihan Liu, Yan Xu, Tiezheng Yu, Wenliang Dai, Ziwei Ji, Samuel Cahyawijaya, Andrea Madotto, and Pascale Fung,
\newblock ``Crossner: Evaluating cross-domain named entity recognition,''
\newblock 2020.

\bibitem{agirre2012semeval}
Eneko Agirre, Daniel Cer, Mona Diab, and Aitor Gonzalez-Agirre,
\newblock ``Semeval-2012 task 6: A pilot on semantic textual similarity. in* sem 2012: The first joint conference on lexical and computational semantics--volume 1: Proceedings of the main conference and the shared task, and volume 2: Proceedings of the sixth international workshop on semantic evaluation (semeval 2012),''
\newblock {\em Association for Computational Linguistics. URL http://www. aclweb. org/anthology/S12-1051}, 2012.

\end{thebibliography}

\end{document}